\RequirePackage{amsmath}
\documentclass[runningheads]{llncs}

\usepackage[utf8]{inputenc} %
\usepackage[T1]{fontenc}    %
\usepackage{hyperref}       %
\usepackage{url}            %
\usepackage{booktabs}       %
\usepackage{amsfonts}       %
\usepackage{nicefrac}       %
\usepackage{microtype}      %
\usepackage{xcolor}         %

\usepackage{cite}

\usepackage{placeins}

\usepackage{amssymb}
\usepackage{mathtools}

\DeclareMathOperator*{\argmax}{argmax}
\DeclareMathOperator*{\argmin}{argmin}
\DeclareMathOperator*{\E}{\mathbb{E}}
\DeclareMathOperator*{\sm}{\mathrm{sm}}

\DeclareMathOperator*{\dec}{\mathrm{dec}}
\newcommand{\cond}{\,|\,}

\usepackage{enumitem}
\setlist{nosep}

\title{Reduction from Complementary-Label Learning to Probability Estimates}

\author{%
  Wei-I Lin%
  \and
  Hsuan-Tien Lin%
}

\institute{%
    National Taiwan University, Taipei, Taiwan \\
    \email{\{r10922076,htlin\}@csie.ntu.edu.tw}
}

\begin{document}

\maketitle

\begin{abstract}

Complementary-Label Learning (CLL) is a weakly-supervised learning problem that aims to learn a multi-class classifier from only complementary labels, which indicate a class to which an instance does not belong. Existing approaches mainly adopt the paradigm of reduction to ordinary classification, which applies specific transformations and surrogate losses to connect CLL back to ordinary classification. Those approaches, however, face several limitations, such as the tendency to overfit. In this paper, we sidestep those limitations with a novel perspective--reduction to probability estimates of complementary classes. We prove that accurate probability estimates of complementary labels lead to good classifiers through a simple decoding step. The proof establishes a reduction framework from CLL to probability estimates. The framework offers explanations of several key CLL approaches as its special cases and allows us to design an improved algorithm that is more robust in noisy environments. The framework also suggests a validation procedure based on the quality of probability estimates, offering a way to validate models with only CLs. The flexible framework opens a wide range of unexplored opportunities in using deep and non-deep models for probability estimates to solve CLL. Empirical experiments further verified the framework's efficacy and robustness in various settings.
\footnote{The full paper can be accessed at \url{https://arxiv.org/abs/2209.09500}.}

\end{abstract}

\section{Introduction}\label{sec:intro}
In real-world machine learning applications, high-quality labels may be hard or costly to collect. To conquer the problem, researchers turn to the \emph{weakly-supervised learning} (WSL) framework, which seeks to learn a good classifier with incomplete, inexact, or inaccurate data \cite{zhou2018brief}.
This paper focuses on a very weak type of WSL, called \emph{complementary-label learning} (CLL) \cite{ishida2017learning}. For the multi-class classification task, a complementary label (CL) designates a class to which a specific instance does not belong. The CLL problem assumes that the learner receives complementary labels rather than ordinary ones during training, while wanting the learner to correctly predict the ordinary labels of the test instances. %
Complementary labels can be cheaper to obtain. For example, when labeling with many classes, selecting the correct label is time-consuming for data annotators, while selecting a complementary label would be less costly \cite{ishida2017learning}. In this case, fundamental studies on CLL models can potentially upgrade multi-class classification models and make machine learning more realistic.
CLL's usefulness also attracts researchers to study its interaction with other tasks, such as generative-discriminative learning \cite{xu2020generative,liu2021gan} and domain-adaptation \cite{zhang2021learning}.

\cite{ishida2017learning,ishida2019complementary} proposed a pioneering model for CLL based on replacing the ordinary classification error with its unbiased risk estimator (URE) computed from only complementary labels assuming that the CLs are generated uniformly. %
\cite{chou2020unbiased} unveiled the overfitting tendency of URE and proposed the surrogate complementary loss (SCL) as an alternative design. %
\cite{yu2018learning} studied the situation where the CLs are not generated uniformly, and proposed a loss function that includes a transition matrix for representing the non-uniform generation. %
\cite{gao2021discriminative} argued that the non-uniform generation shall be tackled by being agnostic to the transition matrix instead of including the matrix in the loss function.

The methods mentioned above mainly focused on applying transformation and specific loss functions to the ordinary classifiers. Such a ``reduction to ordinary classification'' paradigm, however, faces some limitations and is not completely analyzed. For instance, so far most of the methods in the paradigm require differentiable models such as neural networks in their design. It is not clear whether non-deep models could be competitive or even superior to deep ones. 
It remains critical to correct the overfitting tendency caused by the stochastic relationship between complementary and ordinary labels, as repeatedly observed on URE-related methods~\cite{chou2020unbiased}. More studies are also needed to understand the potential of and the sensitivity to the transition matrix in the non-uniform setting, rather than only fixing the matrix in the loss function~\cite{yu2018learning} or dropping it~\cite{gao2021discriminative}.

The potential limitations from reduction to ordinary classification motivate us to sidestep them by taking a different perspective---reduction to complementary probability estimates. Our contribution can be summarized as follows.
\begin{enumerate}
    \item We propose a framework that only relies on the probability estimates of CLs, and prove that a simple decoding method can map those estimates back to correct ordinary labels with theoretical guarantees.
    \item The proposed framework offers explanations of several key CLL approaches as its special cases and allows us to design an improved algorithm that is more robust in noisy environments.
    \item We propose a validation procedure based on the quality of probability estimates, providing a novel approach to validate models with only CLs along with theoretical justifications.
    \item We empirically verify the effectiveness of the proposed framework under broader scenarios than previous works that cover various assumptions on the CL generation (uniform/non-uniform; clean/noisy) and models (deep /non-deep). The proposed framework improves the SOTA methods in those scenarios, demonstrating the effectiveness and robustness of the framework.
\end{enumerate}

\section{Problem Setup}\label{sec:setup}
In this section, we first introduce the problem of ordinary multi-class classification, then formulate the CLL problem, and introduce some common assumption.

\subsection{Ordinary-label learning}
We start by reviewing the problem formulation of ordinary multi-class classification. In this problem, we let $K$ with $K>2$ denote the number of classes to be classified, and use $\mathcal{Y} = [K] = \{1,2,\dotsc,K\}$ to denote the label set. Let $\mathcal{X}\subset\mathbb{R}^d$ denote the feature space. Let $D$ be an unknown joint distribution over $\mathcal{X}\times\mathcal{Y}$ with density function $p_D(x, y)$. Given $N$ i.i.d. training samples $\{(x_i, y_i)\}_{i=1}^N$ and a hypothesis set $\mathcal{H}$, the goal of the learner is to select a classifier $f\colon\mathcal{X}\to\mathbb{R}^K$ from the hypothesis set $\mathcal{H}$ that predicts the correct labels on unseen instances. 
The prediction $\hat{y}$ of an unseen instance $x$ is determined by taking the argmax function on $f$, i.e. $\hat{y} = \argmax_i f_i(x)$, where $f_i(x)$ denote the $i$-th output of $f(x)$. %
The goal of the learner is to learn an $f$ from $\mathcal{H}$ that minimizes the following classification risk:
$\E_{(x,y)\sim D} \big[ \ell(f(x),e_y) \big]$,
where $\ell\colon\mathbb{R}^K\times\mathbb{R}^K \to \mathbb{R}^+$ denotes the loss function, and $e_y$ denote the one-hot vector of label $y$. 

\subsection{Complementary-label learning}\label{subsec:cll}
In complementary-label learning, the goal for the learner remains to find an $f$ that minimizes the ordinary classification risk. 
The difference lies in the dataset to learn from. The complementary learner does not have access to the ground-truth labels $y_i$. 
Instead, for each instance $x_i$, 
the learner is given a complementary label $\bar{y}_i$. A complementary label is a class that $x_i$ does not belong to; that is, $\bar{y}_i \in [K]\backslash\{y_i\}$. In CLL, it is assumed that the complementary dataset is generated according to an unknown distribution $\bar{D}$ over $\mathcal{X}\times\mathcal{Y}$ with density function $\bar{p}_{\bar{D}} (x, y)$. Given access to i.i.d. samples $\{x_i, \bar{y}_i\}_{i=1}^N$ from $\bar{D}$, the complementary-label learner aims to find a hypothesis that classifies the correct ordinary labels on unseen instances. %

Next, we introduce the \emph{class-conditional complementary transition assumption}, which is used by many existing work \cite{ishida2017learning,ishida2019complementary,yu2018learning,gao2021discriminative}. %
It assumes that the generation of complementary labels only depends on the ordinary labels; that is, $P(\bar{y}\cond y,x) = P(\bar{y}\cond y)$. The transition probability $P(\bar{y}\cond y)$ is often represented by a $K\times K$ matrix, called \emph{transition matrix}, with $T_{ij} = P(\bar{y}=j\cond y=i)$. It is commonly assumed to be all-zeros on the diagonals, i.e., $T_{ii} = 0$ for all $i\in[K]$ in CLL because complementary labels are not ordinary.
The transition matrix is further classified into two categories: (a) \emph{Uniform:} In uniform complementary generation, each complementary label is sampled uniformly from all labels except the ordinary one. The transition matrix in this setting is accordingly $T = \frac{1}{K-1}(\mathbf{1}_k - \mathbf{I}_k)$. This is the most widely researched and benchmarked setting in CLL. (b) \emph{Biased:} A biased complementary generation is one that is not uniform. Biased transition matrices could be further classified as invertible ones and noninvertible ones based on its invertibility. The invertibility of a transition matrix comes with less physical meaning in the context of CLL; however, it plays an important role in some theoretical analysis in previous work \cite{yu2018learning,chou2020unbiased}.

Following earlier approaches, we assume that the generation of complementary labels follows class-conditional transition in the rest of the paper and that the transition matrix is given to the learning algorithms. What is different is that we do not assume the transition matrix to be uniform nor invertible. This allows us to make comparison in broader scenarios. In real-world scenario, the true transition matrix may be impossible to access. To loosen the assumption that the true transition matrix is given, we will analyze the case that the given matrix is \emph{inaccurate} later. This analysis can potentially help us understand the CLL in a more realistic environment.

\section{Proposed Framework}\label{sec:framework}
In this section, we propose a framework for CLL based on \emph{complementary probability estimates} (CPE) and \emph{decoding}. We first motivate the proposed CPE framework in Section \ref{subsec:motivation}. Then, we describe the framework and derive its theoretical properties in Section \ref{subsec:cde}. In Section \ref{subsec:connection}, we explain how earlier approaches can be viewed as special cases in CPE. We further draw insights for earlier approaches through CPE and propose improved algorithms based on those insights.

\begin{table*}[t]
\caption{Comparison of recent approaches to CLL. $f(x)$ is the probability estimates of $x$, and $\ell$ is an arbitrary multi-class loss.}
\vskip -0.1in
\label{table:comp-recent}
\centering
\begin{scriptsize}
\begin{tabular}{lcccr}
\toprule
Method & Transformation & Loss Function \\
\midrule
URE \cite{ishida2017learning,ishida2019complementary} &$\phi=I$ &$-(K-1)\ell(f(x),\bar{y})+\sum_{k=1}^K \ell(f(x),k)$\\
SCL-NL \cite{chou2020unbiased} &$\phi = I$ &$-\log(1-f_{\bar{y}}(x))$ \\
Fwd \cite{yu2018learning} &$\phi(f)(x) = T^\top f(x)$ &$\ell(\phi(f)(x), \bar{y})$ \\
DM  \cite{gao2021discriminative} &$\phi(f)(x) = \sm(1 - f(x))$ &$\ell(\phi(f)(x), \bar{y})$ \\
\bottomrule
\end{tabular}
\end{scriptsize}
\vskip -0.25in
\end{table*}
\subsection{Motivation}\label{subsec:motivation}
To conquer CLL, recent approaches \cite{ishida2017learning,yu2018learning,ishida2019complementary,chou2020unbiased,gao2021discriminative} mainly focus on applying different transformation and surrogate loss functions to the ordinary classifier, as summarized in Table \ref{table:comp-recent}. This paradigm of reduction to \emph{ordinary}, however, faces some limitations. For instance, as \cite{chou2020unbiased} points out, the URE approach suffers from the large variance in the gradients. Besides, it remains unclear how some of them behave when the transition matrix is biased. Also, those methods only studied using neural networks and linear models as base models. It is unclear how to easily cast other traditional models for CLL. These limitations motivate us to sidestep them with a different perspective---reduction to \emph{complementary} probability estimates.

\subsection{Methodology}\label{subsec:cde}
\subsubsection{Overview}
The proposed method consists of two steps: In training phase, we aim to find a hypothesis $\bar{f}$ that predicts the distribution of complementary labels well, i.e., an $\bar{f}$ that approximates $P(\bar{y}\cond x)$. This step is motivated by \cite{yu2018learning,gao2021discriminative}, which involve modeling the conditional distribution of the complementary labels $P(\bar{y}\cond x)$, and \cite{zhang2021learning2}, which uses similar idea on noisy-label learning. What is different in our framework is the decoding step during prediction. In inference phase, we propose to predict the label with the closest transition vector to the predicted complementary probability estimates. Specifically, we propose to predict $\hat{y} = \argmin_{k\in[K]} d\left(\bar{f}(x), T_k\right)$ for an unseen instance $x$, where $d$ denotes a loss function. It is a natural choice to decode with respect to $T$ because the transition vector $T_k = (P(\bar{y}=1\cond y=k),\dotsc, P(\bar{y}=K\cond y=k))^\top$ is the ground-truth distribution of the complementary labels if the ordinary label is $k$. In the following paragraph, we provide further details of our framework. %

\subsubsection{Training Phase: Probability Estimates}
In this phase, we aim to find a hypothesis $\bar{f}$ that predicts $P(\bar{y}\cond x)$ well. To do so, given a hypothesis $\bar{f}$ from hypothesis set $\bar{\mathcal{H}}$, we set the following \emph{complementary estimation loss} to optimize:
\begin{equation}
    R(\bar{f};\ell) = \mathbb{E}_{(x,y)\sim\mathcal{D}} \left( \ell(\bar{f}(x), P(\bar{y}\cond x,y)) \right) \label{eq:Rf-origin}
\end{equation}
where $\ell$ can be any loss function defined between discrete probability distributions.
By the assumption that complementary labels are generated with respect to the transition matrix $T$, the ground-truth distribution for $P(\bar{y}\cond x,y)$ is $T_y$, so we can rewrite Equation \eqref{eq:Rf-origin} as follows:
\begin{equation}
    R(\bar{f};\ell) = \mathbb{E}_{(x,y)\sim\mathcal{D}} \left( \ell(\bar{f}(x), T_y) \right) \label{eq:Rf}
\end{equation}
The loss function above is still hard to optimize for two reasons: First, the presence of ordinary label $y$ suggests that it cannot be accessed from the complementary dataset. Second, as we only have \emph{one} complementary label per instance, it becomes questionable to directly use the empirical density, i.e., the one-hot vector of the complementary label $e_{\bar{y}}$ to approximate $T_y$ as it may change the objective.

Here we propose to use the Kullback-Leibler divergence for the loss function to solve the two issues mentioned above with the following property:
\begin{proposition}\label{prop:unbiased}
There is a constant $C$ such that
\begin{equation}
    \E_{(x,\bar{y})\sim\bar{\mathcal{D}}} \ell(\bar{f}(x), e_{\bar{y}}) + C
    = \E_{(x,y)\sim\mathcal{D}} \ell(\bar{f}(x), T_{y})
\end{equation}
holds for all hypothesis $\bar{f}\in\bar{\mathcal{H}}$ if $\ell$ is the KL divergence, i.e., $\ell(\hat{y}, y) = \sum_{k=1}^K -y_k(\log\hat{y}_k - \log y_k)$.
\end{proposition}

The result is well-known in the research of proper scoring rules \cite{kull2015novel,williamson2016composite}. It allows us to replace the $T_y$ by $e_{\bar{y}}$ in Equation \eqref{eq:Rf} because the objective function only differs by a constant after the replacement. This suggests that minimizing the two objectives is equivalent. Moreover, the replacement makes the objective function accessible through the complementary dataset because it only depends on the complementary label $\bar{y}$ rather than the ordinary one.

Formally speaking, minimizing Equation $\eqref{eq:Rf}$ becomes equivalent to minimizing the following \emph{surrogate complementary estimation loss (SCEL)}:
\begin{equation}
    \bar{R}(\bar{f};\ell) = \mathbb{E}_{(x,\bar{y})\sim\bar{\mathcal{D}}} \left( \ell(\bar{f}(x), e_{\bar{y}}) \right) \label{eq:Rbarf}
\end{equation}
By using KL divergence as the loss function, we have that
\begin{equation}
    \bar{R}(\bar{f};\ell) = \mathbb{E}_{(x,\bar{y})\sim\bar{\mathcal{D}}} \left( -\log\bar{f}_{\bar{y}}(x) \right)
\end{equation}
with $\bar{f}_{\bar{y}}(x)$ being the $\bar{y}$-th output of $\bar{f}(x)$. Next, we can use the following empirical version as the training objective:
    $\frac{1}{N}\sum_{i=1}^N -\log\bar{f}_{\bar{y}_i}(x_i)$.
According to the empirical risk minimization (ERM) principle, we can estimate the distribution of complementary labels $P(\bar{y}\cond x)$ by minimizing the log loss on the complementary dataset. That is, by choosing $\bar{f}^*$ with $\bar{f}^* = \argmin_{\bar{f}\in\bar{\mathcal{H}}} \frac{1}{N}\sum_{i=1}^N -\log\bar{f}_{\bar{y}_i}(x_i)$, we can get an estimate of $P(\bar{y}\cond x)$ with $\bar{f}^\ast$.

In essence, we reduce the task of learning from complementary labels into learning probability estimates for multi-class classification (on the \emph{complementary label space}). As the multi-class probability estimates is a well-researched problem, our framework becomes flexible on the choice of the hypothesis set. For instance, one can use K-Nearest Neighbor or Gradient Boosting with log loss to estimate the distribution of complementary labels. %
The flexibility becomes superior to the previous methods, who mainly focus on using neural networks to minimize specific surrogate losses. It makes them hard to optimize for non-differentiable models. In contrast, the proposed methods directly enable existing ordinary models to learn from complementary labels.

\subsubsection{Inference Phase: Decoding}
After finding a complementary probability estimator $\bar{f}^*$ during the training phase, we propose to predict the ordinary label by decoding: Given an unseen example $x$, we predict the label $\hat{y}$ whose transition vector $T_{\hat{y}}$ is closest to the predicted complementary probability estimates. That is, the label is predicted by
\begin{equation}
    \hat{y} = \argmin_{k\in[K]} d\left(\bar{f}^*(x), T_k\right)
\end{equation}
where $d$ could be an arbitrary loss function on the probability simplex and $T_k$ is the $k$-th row vector of $T$. We use $\dec(\bar{f}; d)$ to denote the function that decodes the output from $\bar{f}$ according to the loss function $d$. The next problem is whether the prediction of the decoder can guarantee a small out-sample classification error $R_{01}(f) = \E_{(x,y)\sim\mathcal{D}} I_{f(x)\neq y}$.

We propose to use a simple decoding step by setting $L_1$ distance as the loss function for decoding:
\begin{equation}
    \dec(\bar{f};L_1)\,(x) = \argmin_{y\in[K]} \;\lVert T_y - \bar{f}(x) \rVert_1
\end{equation}
This choice of $L_1$ distance makes the decoding step easy to perform and provides the following bound that quantifies the relationship between the error rate and the quality of probability estimator:
\begin{proposition}\label{prop:ord-comp}
For any $\bar{f}\in\bar{\mathcal{H}}$, and distance function $d$ defined on the probability simplex $\Delta^K$, it holds that
\begin{equation}
    R_{01}\big(\dec(\bar{f};d)\big) \leq \frac{2}{\gamma_d}R(\bar{f}; d)
\end{equation}
where $\gamma_d = \min_{i\neq j} d(T_i, T_j)$ is the minimal distance between any pair of transition vector.
Moreover, if $d$ is the $L_1$ distance and $\ell$ is the KL divergence, then with $\gamma = \min_{i\neq j} \lVert T_i -  T_j\rVert_1$, it holds that
\begin{equation}\label{eq:01-kl-bound}
    R_{01}\big(\dec(\bar{f};L_1)\big) \leq \frac{4\sqrt{2}}{\gamma}\sqrt{R(\bar{f};\ell)}
\end{equation}
\end{proposition}

The proof is in Appendix \ref{sec:proof-ord-comp}. In the realizable case, where there is a target function $g$ that satisfies $g(x) = y$ for all instances, the term $R(\bar{f};\ell_{\text{KL}})$ can be minimized to zero with $\bar{f}^\star: x\mapsto T_{g(x)}$. This indicates that for a sufficiently rich complementary hypothesis set, if the complementary probability estimator is consistent ($\bar{f}\to \bar{f}^\star$) then the $L_1$ decoded prediction is consistent ($R_{01}\big(\dec(\bar{f};L_1)\big)\to 0$). The result suggests that the performance of the $L_1$ decoder can be bounded by the accuracy of the probability estimates of complementary labels measured by the KL divergence. In other words, to obtain an accurate ordinary classifier, it suffices to find an accurate complementary probability estimator followed by the $L_1$ decoding. Admittedly, in the non-realizable case, $R(\bar{f};\ell_{\text{KL}})$ contains irreducible error. We leave the analysis of the error bound in this case for the future research.

Another implication of the Proposition \ref{prop:ord-comp} is related to the inaccurate transition matrix. Suppose the complementary labels are generated with respect to the transition matrix $T'$, which may be different from $T$, the one provided to the learning algorithm. In the proposed framework, the only affected component is the decoding step. This allows us to quantify the effect of inaccuracy as follows:
\begin{corollary}\label{cor:noise}
For any $\bar{f}\in\bar{\mathcal{H}}$, if $d$ is the $L_1$ distance and $\ell$ is the KL divergence, then
\begin{equation}
    R_{01}\big(\dec(f;L_1)\big) \leq \frac{4\sqrt{2}}{\gamma}\sqrt{R(\bar{f};\ell)} + \frac{2\epsilon}{\gamma}.
\end{equation}
where $\gamma = \min_{i\neq j} \lVert T_i -  T_j\rVert_1$ is the minimal $L_1$ distance between pairs of transition vectors, and $\epsilon = \max_{k\in[K]}\lVert T_k' - T_k\rVert_1$ denotes the difference between $T'$ and $T$.
\end{corollary}

\subsubsection{Validation Phase: Quality of Probability Estimates}
The third implication of Proposition \ref{prop:ord-comp} is an alternative validation procedure to the unbiased risk estimation (URE) \cite{ishida2017learning}. According to Proposition \ref{prop:ord-comp}, selecting the best-performing parameter minimizes the right hand side of Eq. \eqref{eq:01-kl-bound} among all hyper-parameter choices minimizes the ordinary classification error. This suggests an alternative metric for parameter selection: using the surrogate complementary estimation loss (SCEL) on the validation dataset.

Although the proposed validation procedure does not directly estimate the ordinary classification error, it provides benefits in the scenarios where URE does not work well. For instance, when the transition matrix is non-invertible, the behavior of URE is ill-defined due to the presence of $T^{-1}$ in the formula of URE: $\E_{x,\bar{y}} e_{\bar{y}}T^{-1}\ell(f(x))$. Indeed, replacing $T^{-1}$ with $T$'s pseudo-inverse can avoid the issue; however, it remains unclear whether the unbiasedness of URE still holds after using pseudo-inverse. In contrast, the quality of complementary probability estimates sidesteps the issue because it does not need to invert the transition matrix. This prevents the proposed procedure from the issue of an ill-conditioned transition matrix.

\subsection{Connection to Previous Methods}\label{subsec:connection}
The proposed framework also explains several earlier approaches as its special cases, including (1) Forward Correction (\textsc{Fwd}) \cite{yu2018learning}, (2) Surrogate Complementary Loss (SCL) with log loss \cite{chou2020unbiased}, and (3) Discriminative Model (DM) \cite{gao2021discriminative}, which are explained in Table~\ref{table:unifiying-view} and Appendix~\ref{sec:equivalence}. By viewing those earlier approaches in the proposed framework, we provide additional benefits for them. First, the novel validation process can be applied for parameter selection. This provides an alternative to validate those approaches. Also, we fill the gap on the theoretical explanation to help understand those approaches in the realizable case.

\begin{table*}[t]
\caption{A unifying view of earlier approaches and proposed algorithms through the lens of reduction to probability estimates, where $U$ denote the uniform transition matrix. Two versions of Forward Correction are considered: General $T$ denotes the original version in \cite{yu2018learning}, and the Uniform denotes the case when the transition layer is fixed to be uniform. Proof of the equivalence is in Appendix \ref{sec:equivalence}.}
\label{table:unifiying-view}
\centering
\vskip -0.1in
\begin{scriptsize}
\begin{tabular}{lcc}
\toprule
Method &Hypothesis set &Decoder \\
\midrule
Fwd (general $T$) \cite{yu2018learning} &$\{x\mapsto T^\top f(x;\theta): \theta\in\Theta\}$ &$\argmax_{k} ((T^\top)^{-1}\bar{f}(x))_k$\\
Fwd (uniform) \cite{yu2018learning}     &$\{x\mapsto U^\top f(x;\theta): \theta\in\Theta\}$ &$\argmin_{k} \lVert \bar{f}(x) - U_k\rVert_1$\\
SCL \cite{chou2020unbiased}          &$\{x\mapsto U^\top f(x;\theta): \theta\in\Theta\}$ &$\argmin_{k} \lVert \bar{f}(x) - U_k\rVert_1$\\ 
DM \cite{gao2021discriminative}         &$\{x\mapsto \sm(1-f(x;\theta)): \theta\in\Theta\}$ &$\argmin_{k} \lVert \bar{f}(x) - U_k\rVert_1$\\
\midrule
CPE-I \scriptsize{(no transition)} &$\{x\mapsto f(x;\theta): \theta\in\Theta\}$        &$\argmin_{k} \lVert \bar{f}(x) - T_k\rVert_1$\\
CPE-F \scriptsize{(fixed transition)} &$\{x\mapsto T^\top f(x;\theta): \theta\in\Theta\}$ &$\argmin_{k} \lVert \bar{f}(x) - T_k\rVert_1$\\
CPE-T \scriptsize{(trainable transition)} &$\{x\mapsto T(W)^\top f(x;\theta): \theta\in\Theta,W\in\mathbb{R}^{K\times K}\}$ &$\argmin_{k} \lVert \bar{f}(x) - T_k\rVert_1$\\
\bottomrule
\end{tabular}
\end{scriptsize}
\vskip -0.25in
\end{table*}

On the other hand, the success of \textsc{Fwd} inspires us to reconsider the role of transition layers in the framework. As the base model's output $f(x;\theta)$ is in the probability simplex $\Delta^K$, the model's output $T^\top f(x;\theta)$ lies in the convex hull formed by the row vectors of $T$. If the transition matrix $T$ provided to the learning algorithm is accurate, then such transformation helps control the model's complexity by restricting its output. The restriction may be wrong, however, when the given transition matrix $T$ is inaccurate. To address this issue, we propose to allow the transition layer to be \emph{trainable}. This technique is also used in label-noise learning, such as \cite{li2021provably}. Specifically, we propose three methods in our \textbf{C}omplementary \textbf{P}robability \textbf{E}stimates framework: (a) \textbf{CPE-I} denotes a model \emph{without} a transition layer (b) \textbf{CPE-F} denotes a model with a \emph{fixed} additional layer to $T$ (c) \textbf{CPE-T} denotes a model with a \emph{trainable} transition layer. To make the transition layer trainable, we considered a $K\times K$ matrix $W$. A softmax function was applied to each row of $W$ to transform it into a valid transition matrix $T(W) = \big(\sm(W_1), \sm(W_2),\dotsc,\sm(W_K)\big)^\top$. For a base model $f$, the complementary probability estimates of \textbf{CPE-T} for a given instance $x$ would be $T(W)^\top f(x;\theta)$. Note that we use the $L_1$ decoder for \textbf{CPE-I}, \textbf{CPE-F}, and \textbf{CPE-T}.

\section{Experiments}\label{sec:experiments}
In this section, we benchmark the proposed framework to the state-of-the-art baselines and discuss the following questions: (a) Can the transition layers improve the model's performance? (b) Is the proposed $L_1$ decoding competitive to \textsc{Max}? (c) Does the transition matrix provide information to the learning algorithms even if it is inaccurate? We further demonstrate the flexibility of incorporating traditional models in \textbf{CPE} in Section~\ref{subsec:trad-models} and verify the effectiveness of the proposed validation procedure in the Appendix.

\subsection{Experiment Setup}\label{subsec:exp-setup}
\subsubsection{Baseline and setup}
We first evaluate CPE with the following state-of-the-art methods: (a) \textbf{URE-GA}: Gradient Ascent applied on the unbiased risk estimator \cite{ishida2017learning,ishida2019complementary}, (b) \textbf{Fwd}: Forward Correction \cite{yu2018learning}, (c) \textbf{SCL}: Surrogate Complementary Loss with negative log loss \cite{chou2020unbiased}, and (d) \textbf{DM}: Discriminative Models with Weighted Loss \cite{gao2021discriminative}.
Following the previous work, we test those methods on MNIST, Fashion-MNIST, and Kuzushiji-MNIST, and use one-layer mlp model (d-500-c) as base models. All models are optimized using Adam with learning rate selected from \{1e-3, 5e-4, 1e-4, 5e-5, 1e-5\} and a fixed weight decay 1e-4 for 300 epochs. The learning rate for \textbf{CPE} is selected with the Surrogate Complementary Estimation Loss (SCEL) on the validation dataset. For the baseline method, it is selected with unbiased risk estimator (URE) of the zero-one loss. It is worth noting that the validation datasets consist of only complementary labels, which is different from some previous works.

\begin{table*}[t]
\caption{Comparison of the testing classification accuracies with different transition matrices (upper part) and different noise levels (lower part).}
\label{table:clean-noise-exp}
\vskip -0.1in
\centering
\begin{tiny}
\begin{tabular}{lrrrrrrrrr}
\toprule
 &\multicolumn{3}{c}{MNIST} &\multicolumn{3}{c}{Fashion-MNIST} &\multicolumn{3}{c}{Kuzushiji-MNIST} \\
 \cmidrule(lr){2-4}
 \cmidrule(lr){5-7}
 \cmidrule(lr){8-10}
 &Unif. &Weak &Strong &Unif. &Weak &Strong &Unif. &Weak &Strong\\
\midrule
URE-GA  &90.3$\pm$ 0.2 &87.8$\pm$ 0.9 &33.8$\pm$ 8.1 &79.4$\pm$ 0.7 &75.7$\pm$ 2.0 &32.3$\pm$ 4.5 &65.6$\pm$ 0.8 &62.5$\pm$ 1.1 &23.3$\pm$ 5.4\\
SCL     &94.3$\pm$ 0.4 &\textbf{93.8$\pm$ 0.4} &27.5$\pm$ 19.8 &82.6$\pm$ 0.4 &81.2$\pm$ 0.1 &28.5$\pm$ 10.8 &\textbf{73.7$\pm$ 1.4} &\textbf{71.2$\pm$ 2.9} &20.7$\pm$ 4.8\\
DM      &91.9$\pm$ 0.6 &90.2$\pm$ 0.3 &26.7$\pm$ 4.6 &82.5$\pm$ 0.3 &80.3$\pm$ 1.1 &24.8$\pm$ 5.0 &65.6$\pm$ 2.9 &64.5$\pm$ 2.7 &20.1$\pm$ 3.2\\
Fwd   &\textbf{94.4$\pm$ 0.2} &91.9$\pm$ 0.3 &95.3$\pm$ 0.4 &82.6$\pm$ 0.6 &\textbf{83.0$\pm$ 1.0} &85.5$\pm$ 0.3 &73.5$\pm$ 1.6 &63.1$\pm$ 2.6 &74.1$\pm$ 4.8\\
\midrule
CPE-I  &90.2$\pm$ 0.2 &88.4$\pm$ 0.3 &92.7$\pm$ 0.8 &81.1$\pm$ 0.3 &79.2$\pm$ 0.5 &81.9$\pm$ 1.4 &66.2$\pm$ 1.0 &62.5$\pm$ 0.9 &73.7$\pm$ 1.0\\
CPE-F  &\textbf{94.4$\pm$ 0.2} &92.0$\pm$ 0.2 &\textbf{95.5$\pm$ 0.3} &\textbf{83.0$\pm$ 0.1} &\textbf{83.0$\pm$ 0.3} &\textbf{85.8$\pm$ 0.3} &73.5$\pm$ 1.6 &64.6$\pm$ 0.5 &\textbf{75.3$\pm$ 2.6}\\
CPE-T  &92.8$\pm$ 0.6 &92.1$\pm$ 0.2 &95.2$\pm$ 0.5 &\textbf{83.0$\pm$ 0.1} &\textbf{83.0$\pm$ 0.3} &\textbf{85.8$\pm$ 0.3} &63.6$\pm$ 0.4 &64.6$\pm$ 0.4 &74.2$\pm$ 2.8\\
\bottomrule
\toprule
\hspace{1em}  &$\lambda=0.1$ &$\lambda=0.2$ &$\lambda=0.5$ &$\lambda=0.1$ &$\lambda=0.2$ &$\lambda=0.5$ &$\lambda=0.1$ &$\lambda=0.2$ &$\lambda=0.5$ \\
\midrule
URE-GA &31.8$\pm$ 6.4 &27.8$\pm$ 8.2 &28.1$\pm$ 4.1 &27.3$\pm$ 5.5 &28.6$\pm$ 4.1 &26.3$\pm$ 2.0 &24.5$\pm$ 4.6 &21.1$\pm$ 2.2 &19.8$\pm$ 2.1\\
SCL &25.1$\pm$ 11.7 &24.7$\pm$ 8.9 &23.8$\pm$ 2.7 &26.6$\pm$ 9.2 &20.6$\pm$ 6.7 &23.2$\pm$ 5.7 &20.4$\pm$ 4.6 &17.3$\pm$ 2.9 &16.8$\pm$ 1.6\\
DM  &26.5$\pm$ 9.1 &24.6$\pm$ 6.5 &22.6$\pm$ 1.3 &24.1$\pm$ 5.1 &23.6$\pm$ 6.7 &22.6$\pm$ 2.9 &20.0$\pm$ 3.0 &19.2$\pm$ 3.1 &18.2$\pm$ 1.6\\
Fwd &88.3$\pm$ 8.7 &83.9$\pm$ 10.7 &71.6$\pm$ 18.4 &\textbf{84.8$\pm$ 0.6} &80.2$\pm$ 6.2 &62.9$\pm$ 20.1 &72.8$\pm$ 5.6 &67.6$\pm$ 7.5 &54.7$\pm$ 12.4\\
\midrule
CPE-I  &92.4$\pm$ 0.7 &92.0$\pm$ 0.8 &87.6$\pm$ 1.4 &81.7$\pm$ 1.4 &81.3$\pm$ 1.4 &78.2$\pm$ 1.5 &73.0$\pm$ 0.7 &71.6$\pm$ 0.9 &62.7$\pm$ 1.6\\
CPE-F  &94.3$\pm$ 0.5 &93.6$\pm$ 0.5 &89.0$\pm$ 1.4 &84.1$\pm$ 0.8 &83.0$\pm$ 1.1 &78.4$\pm$ 2.5 &\textbf{76.1$\pm$ 1.3} &73.7$\pm$ 1.5 &63.7$\pm$ 1.5\\
CPE-T  &\textbf{94.4$\pm$ 0.5} &\textbf{93.7$\pm$ 0.5} &\textbf{89.6$\pm$ 0.9} &84.1$\pm$ 0.8 &\textbf{83.2$\pm$ 1.1} &\textbf{78.9$\pm$ 2.0} &\textbf{76.1$\pm$ 1.3} &\textbf{73.9$\pm$ 1.6} &\textbf{64.2$\pm$ 1.2}\\
\bottomrule
\end{tabular}
\end{tiny}
\end{table*}
\begin{table}[t]
\caption{Comparison of testing accuracies of decoders when the baseline models use fixed transition layers. The parameters are selected from the one with smallest SCEL on the validation dataset.}
\label{table:bench-exp-cpe}
\centering
\vskip -0.1in
\begin{tiny}
\begin{tabular}{lrrrrrrrrr}
\toprule
&\multicolumn{3}{c}{MNIST} &\multicolumn{3}{c}{Fashion-MNIST} &\multicolumn{3}{c}{Kuzushiji-MNIST} \\
 \cmidrule(lr){2-4}
 \cmidrule(lr){5-7}
 \cmidrule(lr){8-10}
&Unif. &Weak &Strong &Unif. &Weak &Strong &Unif. &Weak &Strong\\
\midrule
\textsc{Max}  &94.4$\pm$ 0.2 &92.0$\pm$ 0.2 &95.5$\pm$ 0.2 &83.0$\pm$ 0.1 &\textbf{83.3$\pm$ 0.2} &\textbf{86.1$\pm$ 0.5} &73.5$\pm$ 1.6 &\textbf{64.8$\pm$ 0.5} &75.3$\pm$ 2.6\\
$L_1$  &94.4$\pm$ 0.2 &92.0$\pm$ 0.2 &95.5$\pm$ 0.3 &83.0$\pm$ 0.1 &83.0$\pm$ 0.3 &85.8$\pm$ 0.3 &73.5$\pm$ 1.6 &64.6$\pm$ 0.5 &75.3$\pm$ 2.6\\
\bottomrule
\toprule
&$\lambda=0.1$ &$\lambda=0.2$ &$\lambda=0.5$ &$\lambda=0.1$ &$\lambda=0.2$ &$\lambda=0.5$ &$\lambda=0.1$ &$\lambda=0.2$ &$\lambda=0.5$ \\
\midrule
\textsc{Max}  &\textbf{94.4$\pm$ 0.3} &93.5$\pm$ 0.3 &84.5$\pm$ 4.1 &\textbf{85.0$\pm$ 0.3} &\textbf{84.0$\pm$ 0.5} &76.5$\pm$ 2.5 &\textbf{76.4$\pm$ 1.1} &\textbf{73.8$\pm$ 1.2} &59.9$\pm$ 3.4\\
$L_1$  &94.3$\pm$ 0.5 &\textbf{93.6$\pm$ 0.5} &\textbf{89.0$\pm$ 1.4} &84.1$\pm$ 0.8 &83.0$\pm$ 1.1 &\textbf{78.4$\pm$ 2.5} &76.1$\pm$ 1.3 &73.7$\pm$ 1.5 &\textbf{63.7$\pm$ 1.5}\\
\bottomrule
\end{tabular}
\end{tiny}
\end{table}
\begin{table*}[t]
\caption{Comparison of testing accuracies of CPE with traditional models. \textbf{Boldfaced} ones outperform the baseline methods based on single-layer deep models.}
\label{table:trad-bench-exp}
\vskip -0.1in
\centering
\begin{tiny}
\begin{tabular}{lrrrrrrrrr}
\toprule
 &\multicolumn{3}{c}{MNIST} &\multicolumn{3}{c}{Fashion-MNIST} &\multicolumn{3}{c}{Kuzushiji-MNIST} \\
 \cmidrule(lr){2-4}
 \cmidrule(lr){5-7}
 \cmidrule(lr){8-10}
Model &Unif. &Weak &Strong &Unif. &Weak &Strong &Unif. &Weak &Strong \\
\midrule
CPE-KNN    &93.1$\pm$ 0.1 &92.6$\pm$ 0.1 &94.5$\pm$ 0.4 &79.1$\pm$ 0.4 &77.8$\pm$ 0.6 &79.0$\pm$ 1.7 &\textbf{74.9$\pm$ 0.8} &\textbf{73.7$\pm$ 0.8} &\textbf{80.4$\pm$ 1.3}\\
CPE-GBDT   &86.9$\pm$ 0.4 &86.0$\pm$ 0.3 &90.3$\pm$ 0.9 &79.8$\pm$ 0.4 &78.0$\pm$ 0.4 &81.4$\pm$ 1.1 &60.6$\pm$ 0.4 &56.6$\pm$ 1.8 &68.4$\pm$ 2.1\\
\bottomrule
\toprule
  &$\lambda=0.1$ &$\lambda=0.2$ &$\lambda=0.5$ &$\lambda=0.1$ &$\lambda=0.2$ &$\lambda=0.5$ &$\lambda=0.1$ &$\lambda=0.2$ &$\lambda=0.5$ \\
\midrule
CPE-KNN   &93.7$\pm$ 0.4 &93.4$\pm$ 0.4 &\textbf{91.9$\pm$ 1.1} &78.7$\pm$ 1.9 &78.5$\pm$ 1.9 &76.6$\pm$ 1.9 &\textbf{77.2$\pm$ 1.1} &\textbf{75.9$\pm$ 1.6} &\textbf{73.2$\pm$ 1.7}\\
CPE-GBDT  &89.7$\pm$ 1.0 &88.6$\pm$ 1.2 &84.0$\pm$ 1.7 &80.6$\pm$ 1.7 &80.0$\pm$ 1.6 &76.0$\pm$ 2.2 &66.7$\pm$ 2.4 &64.7$\pm$ 2.4 &55.8$\pm$ 3.1\\
\bottomrule
\end{tabular}
\end{tiny}
\vskip -0.2in
\end{table*}

\subsubsection{Transition matrices}
In the experiment of \emph{clean} transition matrices, three types of transition matrices are benchmarked in the experiment. Besides the uniform transition matrix, following \cite{yu2018learning,gao2021discriminative}, we generated two biased ones as follows: For each class $y$, the complementary classes $\mathcal{Y}\backslash\{y\}$ are first randomly split into three subsets. Within each subset, the probabilities are set to $p_1$, $p_2$ and $p_3$, respectively. We consider two cases for $(p_1, p_2, p_3)$: (a) \emph{Strong}: $(\frac{0.75}{3}, \frac{0.24}{3}, \frac{0.01}{3})$ to model stronger deviation from uniform transition matrices. (b) \emph{Weak}: $(\frac{0.45}{3}, \frac{0.30}{3}, \frac{0.25}{3})$ to model milder deviation from uniform transition matrices.
In the experiment of \emph{noisy} transition matrices, we consider the \emph{Strong} deviation transition matrix $T_\text{strong}$ to be the ground-truth transition matrix, and a uniform noise transition matrix $\frac{1}{K}\mathbf{1}_K$ to model the noisy complementary label generation. We generated complementary labels with the transition matrix $(1-\lambda) T_\text{strong} + \lambda \frac{1}{K}\mathbf{1}_K$, but provided $T_\text{strong}$ and the generated complementary dataset to the learners. The parameter $\lambda$ controls the proportion of the uniform noise in the complementary labels. The results are reported in Table~\ref{table:clean-noise-exp}.

\subsection{Discussion}

\subsubsection{Can Transition Layers Improve Performance?}
The answer is positive in both clean and noisy experiments. We observe that \textbf{CPE-F} and \textbf{CPE-T} outperform \textbf{CPE-I} in both settings, demonstrating that the transition layer help achieve higher performances, no matter the provided transition matrix is clean or not. Also, we observe that \textbf{CPE-T} outperforms \textbf{CPE-F} in the noisy setting, especially when the noise factor $\lambda$ is large. It demonstrates that by making transition layers trainable, the model can potentially fit the distribution of complementary labels better by altering the transition layer. In contrast, \textbf{CPE-F} is restricted to a wrong output space, making it underperform \textbf{CPE-T}. The difference makes \textbf{CPE-T} a better choice for noisy environment.

\subsubsection{Is $L_1$ competitive with \textsc{Max}?}
As analyzed in Section \ref{subsec:connection}, \textbf{Fwd} and \textbf{CPE-F} only differ in the decoding step, with the former using \textsc{Max} and the latter using $L_1$. We provide the testing accuracies of these decoders when the base models are \textbf{CPE-F} in Table \ref{table:bench-exp-cpe}. It is displayed that the \textsc{Max} decoder outperform $L_1$ in most noiseless settings; however, when the transition matrix is highly inaccurate ($\lambda=0.5$), we observe that the $L_1$ decoder outperform the \textsc{Max} decoder. This suggests that $L_1$ could be more tolerant to an inaccurate transition matrix. These results reveal that a deeper sensitivity analysis of different decoders, both empirically and theoretically, would be desired. We leave this as future studies.

\subsubsection{Discussion of $T$-agnostic models}
Among the baseline methods, \textbf{URE-GA}, \textbf{SCL} and \textbf{DM} are ones that does not take $T$ as inputs or assumes $T$ is uniform, which we called $T$-agnostic models. Those models perform well when the transition matrix is just slightly deviated from the uniform one, but their performances all dropped when the deviation from uniform becomes larger. As we discussed in Section \ref{subsec:connection}, the result can be interpreted to be caused by their implicit assumption on uniform transition matrices, which brings great performance on uniform transition matrices but worse performance on biased ones.
In contrast, we observed that all variations of \textbf{CPE} have similar testing accuracies across different transition matrices, demonstrating that \textbf{CPE} does exploit the information from the transition matrix that helps the models deliver better performance.

\subsection{Learn from CL with Traditional Methods} \label{subsec:trad-models}
As discussed in Section \ref{sec:framework}, the proposed framework is not constrained by deep models. We explored the possibility of applying traditional methods to learn from CL, including (a) $k$-Nearest Neighbor (\textbf{$k$-NN}) and (b) Gradient Boosting Decision Tree (\textbf{GBDT}). We benchmarked those models in the same settings and reported the restuls in Table \ref{table:trad-bench-exp}. It displays that traditional models, specifically, \textbf{$k$-NN}, outperform all the methods using deep models in Kuzushiji-MNIST, indicating the benefit of the proposed CPE's flexibility in using non-deep models.

\section{Conclusion}\label{sec:conclusion}
In this paper, we view the CLL problem from a novel perspective, reduction to complementary probability estimates. Through this perspective, we propose a framework that only requires complementary probability estimates and prove that a simple decoding step can map the estimates to ordinary labels. The framework comes with a theoretically justified validation procedure, provable tolerance in noisy environment, and flexibility of incorporating non-deep models. Empirical experiments further verify the effectiveness and robustness of the proposed framework under broader scenarios, including non-uniform and noisy complementary label generation. We expect the realistic elements of the framework to keep inspiring future research towards making CLL practical.

\bibliography{paper}
\bibliographystyle{splncs04}

\subsubsection*{Acknowlegements.}
We thank the anonymous reviewers and the members of NTU CLLab for valuable suggestions.
The work is partially supported by the National Science and Technology Council via the grants 110-2628-E-002-013 and 111-2628-E-002-018.
We also thank the National Center for High-performance Computing (NCHC) of National Applied Research Laboratories (NARLabs) in Taiwan for providing computational resources.

\appendix
\section{Proofs} \label{sec:proofs}
This section provides the proofs for the propositions, theorems claimed in the main text.
\subsection{Proof of Proposition \ref{prop:unbiased}} \label{sec:proof-unbiased}
First, set $C = \E_{(x,y)\sim\mathcal{D}} \sum_{k=1}^K T_{yk}\log(T_{yk})$, then
\begin{equation}
    \E_{(x,y)\sim\mathcal{D}} \ell(\bar{f}(x), T_{y})
    = \E_{(x,y)\sim\mathcal{D}} \sum_{k=1}^K -T_{yk}\log\left(\frac{\bar{f}_k(x)}{T_{yk}}\right)
    = C + \E_{(x,y)\sim\mathcal{D}} \sum_{k=1}^K -T_{yk}\log(\bar{f}_k(x))
\end{equation}
Next, as $P(\bar{y}\cond y) = T_{y\bar{y}}$, then 
\begin{equation}
    \E_{(x,y)\sim\mathcal{D}} \sum_{k=1}^K -T_{yk}\log(\bar{f}_k(x))
    = \E_{(x,y)\sim\mathcal{D}} \left(\E_{\bar{y}\cond y} -\log(\bar{f}_{\bar{y}}(x))\right)
    = \E_{(x,\bar{y})\sim\bar{\mathcal{D}}} \ell(\bar{f}(x), e_{\bar{y}})
\end{equation}
Hence, $\E_{(x,y)\sim\mathcal{D}} \ell(\bar{f}(x), T_{y}) = C +  \E_{(x,\bar{y})\sim\bar{\mathcal{D}}} \ell(\bar{f}(x), e_{\bar{y}})$.

\subsection{Proof of Proposition \ref{prop:ord-comp}} \label{sec:proof-ord-comp}
Let $I_A$ denote the indicator function of event $A$, then using Markov's inequality on the random variable $d(\bar{f}(x), T_y)$, we have
\begin{equation}\label{eq:markov}
    R_{01}\big(\dec(\bar{f};d)\big)
    \leq P\Big(d(\bar{f}(x), T_y)\geq \frac{\gamma_d}{2} \Big)
    \leq \frac{2}{\gamma_d} \E \Big[d(\bar{f}(x), T_y)\Big]
    = \frac{2}{\gamma_d} R(\bar{f};d)
\end{equation}
To see the first inequality holds, note that if $d(\bar{f}(x), T_y) < \frac{\gamma_d}{2}$, then for any incorrect class $y'\neq y$, we have
\begin{equation}
    d(\bar{f}(x), T_{y'}) \geq d(T_y, T_{y'}) - d(T_y, \bar{f}(x)) \geq \frac{\gamma_d}{2}
\end{equation}
by triangular inequality and the definition of $\gamma_d$. As a result, the decoder decodes $\bar{f}(x)$ to the correct class $y$ if $d(\bar{f}(x), T_y) < \frac{\gamma_d}{2}$. This completes the first part of the Proposition.

Next, by Pinsker's inequality and Jensen's inequality, we have that
\begin{align}\label{eq:l1-kl}
    R(\bar{f};L_1) &= \E_{(x,y)\sim\mathcal{D}} \big\lVert \bar{f}(x) - T_y \big\rVert_1  \\
        &\leq 2\E_{(x,y)\sim\mathcal{D}} \sqrt{2\ell_\text{KL} \big( \bar{f}(x), T_y \big)} \\
        &\leq 2\sqrt{2\E_{(x,y)\sim\mathcal{D}} \ell_\text{KL}\big( \bar{f}(x), T_y \big) }
        = 2\sqrt{2R(\bar{f};\ell_\text{KL})}
\end{align}
According to the above inequality and the results of the first part, the proof for the second part is now complete.

\subsection{Proof of Corollary \ref{cor:noise}} \label{sec:proof-noise}
The decoding step remains the same when $T'\neq T$ because the decoder uses the same transition matrix $T$ to decode. The only difference is in the complementary probability estimates. Specifically, we have that the complementary estimation loss becomes $R(\bar{f}; \ell) = \mathbb{E}_{(x,y)\sim\mathcal{D}} \left( \ell(\bar{f}(x), T'_y) \right)$ as the complementary labels are generated with respect to $T'$.

Hence, the last equality in Equation \eqref{eq:markov} is no longer correct. Instead, we use the following:
\begin{equation}
    \E \Big[d(\bar{f}(x), T_y)\Big]
    \leq \E \Big[d(\bar{f}(x), T'_y) + d(T'_y, T_y)\Big]
    \leq \E \Big[d(\bar{f}(x), T'_y)\Big] + \epsilon
\end{equation}
to obtain that $R_{01}\big(\dec(\bar{f};d)\big) \leq \frac{2}{\gamma_d} R(\bar{f};d) + \frac{2\epsilon}{\gamma_d}$. Then, we can use Pinsker's inequality and Jensen's inequality as in \eqref{eq:l1-kl} to get
\begin{equation}
    R_{01}\big(\dec(f;L_1)\big) \leq \frac{4\sqrt{2}}{\gamma}\sqrt{R(\bar{f};\ell)} + \frac{2\epsilon}{\gamma}.
\end{equation}

\section{Details of the Connections between Proposed Framework and Previous Methods}\label{sec:equivalence}
In this section, we provide further details about how our framework can explain several previous methods as its special cases. Across this section, we let $f(\cdot;\theta)$ denote the base model parametrized by $\theta\in\Theta$. We also provide some insights drawn from viewing these previous methods using the proposed framework.

\paragraph{Forward Correction}
In the training phase, Forward Correction optimizes the following loss functions:
\begin{equation} \label{eq:fwd-loss}
    L_\text{Fwd}(\theta) = \frac{1}{N}\sum_{i=1}^N -\log\big(T^\top f(x_i;\theta)\big)_{\bar{y}_i}
\end{equation}
In the inference phase, Forward Correction predicts $\hat{y} = \argmax_k f_k(x)$ for an unseen instance $x$. We claim that Forward Correction is equivalent to CPE with the following parameters when $T$ is invertible:
\begin{itemize}
    \item Hypothesis Set: $\{x\mapsto T^\top f(x;\theta): \theta\in\Theta \}$
    \item Decoder: $\argmax_{k} \big((T^\top)^{-1}\bar{f}(x;\theta)\big)_k$.
\end{itemize}
\begin{proof}
First, by setting the hypothesis set as above and plugging in the surrogate complementary estimation loss, we get the training objective function for CPE:
\begin{equation}\label{eq:cpe-fwd}
    L_\text{CPE}(\theta) = \frac{1}{N}\sum_{i=1}^N -\log \big(T^\top f(x_i;\theta)\big)_{\bar{y}_i}
\end{equation}
Equation \eqref{eq:cpe-fwd} matches Equation \eqref{eq:fwd-loss}, implying that in the training phase they select the same parameter $\theta$. Next, in the inference phase, it is clear that $(T^\top)^{-1} \bar{f}(x;\theta) = (T^\top)^{-1}T^\top f(x;\theta) = f(x;\theta)$, so both methods predict the same label for an instance $x$.
\end{proof}

Next, we further show that when $T$ is the uniform transition matrix $U$, the decoder is equivalent to the $L_1$ decoder, i.e., $\argmax_{k} ((U^\top)^{-1}\bar{f}(x))_k = \argmin_{k} \lVert U_k - \bar{f}(x)\rVert_1$:
\begin{proof}
First, as
\[
    ((U^\top)^{-1}\bar{f}(x))_k = -(K-1)\bar{f}_k(x) + \sum_{k=1}^K \bar{f}_k(x) = - (K-1)\bar{f}_k(x) + 1,
\]
we have that $\argmax_{k} ((U^\top)^{-1}\bar{f}(x))_k = \argmin_{k} \bar{f}_k(x)$. Next, set $\hat{y} = \argmin_k \bar{f}_k(x)$. For any $y\neq \hat{y}$, we want to show
\begin{equation} \label{eq:goal}
    |U_{y\hat{y}}  - \bar{f}_{\hat{y}}(x)| + |U_{yy} - \bar{f}_y(x)| \geq |U_{\hat{y}\hat{y}}  - \bar{f}_{\hat{y}}(x)| + |U_{\hat{y}y} - \bar{f}_y(x)|.
\end{equation}
As $\bar{f}_{\hat{y}}(x)\leq \frac{1}{K} \leq \frac{1}{K-1} = U_{y\hat{y}}$,
\begin{align}
    |U_{y\hat{y}} - \bar{f}_{\hat{y}}(x)| + |U_{yy} - \bar{f}_y(x)|
    &= |U_{y\hat{y}} - \bar{f}_{\hat{y}}(x)| + \bar{f}_{\hat{y}}(x) + |U_{yy} - \bar{f}_y(x)| - f_{\hat{y}}(x) \\
    &= |U_{\hat{y}\hat{y}} - \bar{f}_{\hat{y}}(x)| + |U_{y\hat{y}} - \bar{f}_{\hat{y}}(x)| + |U_{yy} - \bar{f}_y(x)| - \bar{f}_{\hat{y}}(x) \\
    &= |U_{\hat{y}\hat{y}} - \bar{f}_{\hat{y}}(x)| + \frac{1}{K-1} - \bar{f}_{\hat{y}}(x) + \bar{f}_y(x) - \bar{f}_{\hat{y}}(x)
\end{align}
If $\bar{f}_y(x) \leq \frac{1}{K-1}$, as $\bar{f}_{\hat{y}}(x) \leq \bar{f}_y(x)$,
\[
    \frac{1}{K-1} - \bar{f}_{\hat{y}}(x) + \bar{f}_y(x) - \bar{f}_{\hat{y}}(x)
    \geq \frac{1}{K-1} - \bar{f}_{\hat{y}}(x)
    \geq \frac{1}{K-1} - \bar{f}_{y}(x)
    = |U_{\hat{y}y} - \bar{f}_y(x)|
\]
Otherwise, as $\bar{f}_{\hat{y}}(x) \leq \frac{1}{K}$,
\[
    \frac{1}{K-1} - \bar{f}_{\hat{y}}(x) + \bar{f}_y(x) - \bar{f}_{\hat{y}}(x)
    \geq \bar{f}_y(x) - \bar{f}_{\hat{y}}(x)
    \geq \frac{1}{K-1} - \bar{f}_{y}(x)
    = |U_{\hat{y}y} - \bar{f}_y(x)|.
\]
Hence, Equation \eqref{eq:goal} holds. Now,
\begin{align}
\sum_{k=1}^K \left| U_{yk} - \bar{f}_k(x) \right|
    &= \left| U_{y\hat{y}} - \bar{f}_{\hat{y}}(x) \right| + \left| U_{yy} - \bar{f}_{y}(x) \right| + \sum_{k\neq y,\hat{y}}\left| U_{yk} - \bar{f}_k(x) \right| \\ \label{eq:eq-0}
    &\geq \left| U_{\hat{y}y} - \bar{f}_{y}(x) \right| + \left| U_{\hat{y}\hat{y}} - \bar{f}_{\hat{y}}(x) \right| + \sum_{k\neq y,\hat{y}}\left| U_{\hat{y}k} - \bar{f}_k(x) \right|
    = \sum_{k=1}^K \left| U_{\hat{y}k} - \bar{f}_k(x) \right|
\end{align}
As a result, $\hat{y}$ minimizes $k\mapsto \lVert U_k - \bar{f}(x)\rVert_1$. Hence, we conclude that $\argmin_{k} \bar{f}_k(x) = \bar{y} = \argmin_{k} \lVert U_k - \bar{f}_k(x)\rVert_1$. Then the proof is complete.
\end{proof}
As the two decoders are equivalent, we have that Forward Correction is equivalent to CPE with
\begin{itemize}
    \item Hypothesis Set: $\{x\mapsto U^\top f(x;\theta): \theta\in\Theta \}$
    \item Decoder: $\argmin_{k} \lVert \bar{f}(x;\theta) - U_k\rVert_1 $.
\end{itemize}
when the transition layer is fixed to the uniform transition matrix.

\paragraph{Surrogate Complementary Loss}
In the training phase, Surrogate Complementary Loss with Log Loss optimizes the following loss functions:
\begin{equation}
    L_\text{SCL}(\theta) = \frac{1}{N}\sum_{i=1}^N -\log (1 - f(x_i;\theta))_{\bar{y}_i}
\end{equation}
In the inference phase, this method predicts the ordinary labels by $\hat{y} = \argmax_k f_k(x)$ for an unseen instance $x$. We claim that this method is equivalent CPE with:
\begin{itemize}
    \item Hypothesis Set: $\{x\mapsto U^\top f(x;\theta): \theta\in\Theta \}$
    \item Decoder: $\argmin_{k} \lVert \bar{f}(x;\theta) - U_k\rVert_1 $.
\end{itemize}
\begin{proof}
Observe that the training objective function for CPE with the hypothesis set has the following property:
\begin{align}
L_\text{CPE}(\theta) 
    &= \frac{1}{N}\sum_{i=1}^N -\log \left(U^\top f(x_i;\theta)_{\bar{y}_i}\right) 
    = \frac{1}{N}\sum_{i=1}^N -\log \Bigg(\frac{1}{K-1}\sum_{k\neq \bar{y}_i} f_k(x_i;\theta)\Bigg)\\
    &= \frac{1}{N}\sum_{i=1}^N -\log \big(1 - f_{\bar{y}_i}(x_i;\theta)\big) + \log(K-1) 
    = L_\text{SCL}(\theta) + \log(K-1)
\end{align}
That is, the objective function only differs by a constant. As a result, the two methods match during the training phase.

In inference phase, SCL predicts $\hat{y} = \argmax_{k} f(x;\theta)$ for unseen instance $x$ as in Forward Correction. In addition, they have the same hypothesis set $\{x\mapsto U^\top f(x;\theta): \theta\in\Theta \}$ if the transition layer of Forward Correction is fixed to uniform. Hence, SCL is equivalent to Forward Correction with uniform transition layer. It implies that they have the same decoder: $\hat{y} = \argmin_{k} \lVert \bar{f}(x) - U_k\rVert_1$.
\end{proof}

\paragraph{Discriminative Model}
In the training phase, Discriminative Model with unweighted loss optimizes the following loss functions:
\begin{equation}
    L_\text{DM}(\theta) = \frac{1}{N}\sum_{i=1}^N -\log \big(\sm(1-f(x_i;\theta))\big)_{\bar{y}_i}
\end{equation}
In the inference phase, this method predicts the ordinary labels by $\hat{y} = \argmax_k f_k(x)$ for an unseen instance $x$. We claim that this method is equivalent CPE with:
\begin{itemize}
   \item Hypothesis Set: $\{x\mapsto \sm(1-f(x;\theta)): \theta\in\Theta \}$
   \item Decoder: $\argmin_{k} \lVert \bar{f}(x;\theta) - U_k\rVert_1 $.
\end{itemize}
\begin{proof}
The equivalence in the training phase is clear by plugging in the hypothesis to the surrogate complementary estimation loss. During inference phase, first observe that
\begin{equation}
    \bar{f}_k(x) = \frac{1}{Z} \exp\big(1-f_k(x_i;\theta)\big) = \frac{e}{Z} \exp\big(-f_k(x_i;\theta)\big),
\end{equation}
where $Z = \sum_{k=1}^K \exp\big(1-f_k(x_i;\theta)\big)$ is the normalization term. As $x\mapsto \exp(-x)$ is monotonic decreasing, we have that $\argmin_k \bar{f}_k(x;\theta) = \argmax_k f_k(x;\theta)$. Next, as we have shwon $\argmin_{k} \bar{f}_k(x) = \argmin_{k} \lVert U_k - \bar{f}_k(x)\rVert_1$, so $\argmax_k f_k(x;\theta) = \argmin_{k} \lVert U_k - \bar{f}_k(x)\rVert_1$, implying that both methods predict the same label for all instances.
\end{proof}

\paragraph{Observations by viewing earlier approaches with the proposed framework}
We also draw the following observations by viewing earlier approaches with the proposed CPE framework:
\begin{enumerate}
\item By viewing \textsc{Fwd} with the proposed framework, the equivalent decoder essentially converts the complementary probability estimates back to the ordinary probability estimates and predicts the largest one. We name it \textsc{Max} decoding for future reference. 
\item If the transition matrix is uniform, then \textsc{Fwd} and \textsc{SCL} with log loss match, suggesting that they are the same in this situation. It explains why those two methods have similar performances in \cite{chou2020unbiased}, which is also reproduced in our experiment, reported in Table \ref{table:clean-noise-exp}.
\item DM was proposed to lift the generation assumption of complementary labels \cite{gao2021discriminative}, but from the view of the CPE framework, DM implicitly assumes the complementary labels are generated uniformly, as we can see from the decoder. This provides an alternative explanation why its performance deteriorates as the transition matrix deviates from the uniform matrix, as shown in \cite{gao2021discriminative}.
\end{enumerate}

\newpage
\section{Experiment Details} \label{sec:details}
In this section, we provide missing details of the experiments in Section \ref{sec:experiments}.
\subsection{Setup}
\paragraph{Datasets}
Across the experiments, we use the following datasets:
\begin{itemize}
    \item MNIST %
    \item Fashion-MNIST %
    \item Kuzushiji-MNIST %
\end{itemize}
For the above dataset, the size of the training set is 60000, and the size of the testing set is 10000. To perform the hyperparameter selection, in each trial, we split 10 percent of the training dataset randomly as the validation dataset. We performed five trials with different random seeds for all the experiments in this paper. To ensure a fair comparison, the dataset split and the generated complementary labels are the same for the benchmark algorithms. Also, we did not include data augmentation or consistency regularization \cite{wang2021learning} in the experiment to prevent introducing extra factors and simplify the comparison.

\paragraph{Models}
We implemented the deep models in PyTorch. The base models considered in the experiment are linear and one-layer mlp model (d-500-c) with 500 hidden units. In CPE-T, the parameter of the transition layer is initialized such that it matches the provided transition matrix, i.e. it is initialized to $W_0$ such that $T(W_0)=T$. All models are optimized using Adam with learning rate selected from \{1e-3, 5e-4, 1e-4, 5e-5, 1e-5\} and a fixed weight decay 1e-4 for 300 epochs. We used the default parameters in PyTorch for other parameters in Adam. The experiments are run with Nvidia Tesla V100 GPUs.

For the two traditional models, we used the K nearest neighbor (KNN) classifier from scikit-learn with the number of neighbors selected from $\{10,20,\dotsc,250\}$ based on the complementary estimation loss on the validation dataset. We performed PCA on the dataset to map the feature to a $32$-dimension space for KNN to reduce the training/inference time. We used Gradient Boosting Decision Tree from LightGBM, and set the objective to ``multiclass'' to optimize the log loss. The hyperparameters include the number of trees $\{5,10,\dotsc,500\}$ and learning rate $\{0.01, 0.025, 0.05, 0.1\}$. Those parameters are also selected based on the complementary estimation loss on the validation dataset.

\subsection{Additional Results} \label{subsec:additional-results}
This section provides figures and tables that are helpful in analyzing the experiment results.

\newpage
\paragraph{Benchmark results of linear models}
Table \ref{table:bench-exp-full} and \ref{table:noise-exp-full} provide the the noiseless and noisy benchmark results using linear models as base models, using the same setting in Section~\ref{subsec:exp-setup}. We can see that the proposed CPE performs slightly better or is competitive with the baseline methods in most scenarios. When the transition matrix is highly inaccurate ($\lambda=0.5$), CPE outperforms the baselines and is more stable in terms of testing accuracies. These are consistent with our observation when using mlp as base models.

\begin{table*}[t]
\caption{Comparison of the testing classification accuracies with different transition matrices.}
\vskip -0.2in
\label{table:bench-exp-full}
\begin{center}
\begin{tiny}
\begin{sc}
\begin{tabular}{lrrrrrrrrr}
\toprule
 &\multicolumn{3}{c}{MNIST} &\multicolumn{3}{c}{Fashion-MNIST} &\multicolumn{3}{c}{Kuzushiji-MNIST} \\
 \cmidrule(lr){2-4}
 \cmidrule(lr){5-7}
 \cmidrule(lr){8-10}
 &Unif. &Weak &Strong &Unif. &Weak &Strong &Unif. &Weak &Strong\\
\midrule
URE-GA &81.7$\pm$ 0.5 &73.4$\pm$ 1.4 &23.7$\pm$ 2.9 &76.2$\pm$ 0.3 &70.8$\pm$ 1.5 &21.3$\pm$ 5.5 &51.0$\pm$ 1.0 &43.7$\pm$ 1.0 &16.7$\pm$ 2.5\\
SCL    &\textbf{90.5$\pm$ 0.2} &90.2$\pm$ 0.2 &25.0$\pm$ 17.9 &82.0$\pm$ 0.4 &79.6$\pm$ 2.2 &26.2$\pm$ 8.7 &59.9$\pm$ 0.9 &58.9$\pm$ 0.7 &16.4$\pm$ 2.2\\
DM     &89.7$\pm$ 0.5 &89.1$\pm$ 0.2 &22.7$\pm$ 8.5 &81.8$\pm$ 0.3 &78.2$\pm$ 3.1 &23.6$\pm$ 5.5 &\textbf{61.0$\pm$ 1.5} &59.4$\pm$ 1.4 &17.7$\pm$ 3.0\\
Fwd    &\textbf{90.5$\pm$ 0.2} &90.6$\pm$ 0.4 &91.6$\pm$ 0.7 &82.0$\pm$ 0.4 &81.6$\pm$ 1.2 &\textbf{83.4$\pm$ 0.7} &59.9$\pm$ 0.9 &60.4$\pm$ 0.9 &62.6$\pm$ 0.7\\
CPE-I  &80.4$\pm$ 0.3 &73.5$\pm$ 1.3 &76.1$\pm$ 1.6 &74.6$\pm$ 0.5 &71.0$\pm$ 1.5 &74.7$\pm$ 2.3 &49.7$\pm$ 0.6 &42.8$\pm$ 0.8 &46.8$\pm$ 1.4\\
CPE-F  &\textbf{90.5$\pm$ 0.2} &\textbf{90.7$\pm$ 0.1} &\textbf{91.8$\pm$ 0.4} &\textbf{82.2$\pm$ 0.3} &\textbf{82.4$\pm$ 0.4} &83.1$\pm$ 1.0 &60.4$\pm$ 0.6 &\textbf{60.8$\pm$ 0.4} &62.8$\pm$ 0.2\\
CPE-T  &\textbf{90.5$\pm$ 0.2} &90.6$\pm$ 0.1 &\textbf{91.8$\pm$ 0.4} &82.0$\pm$ 0.3 &82.1$\pm$ 0.5 &83.2$\pm$ 1.2 &60.3$\pm$ 0.5 &60.6$\pm$ 0.5 &\textbf{63.0$\pm$ 0.3}\\
\bottomrule
\end{tabular}
\end{sc}
\end{tiny}
\end{center}
\end{table*}
\begin{table*}[t]
\caption{Comparison of the testing classification accuracies with different levels of noise.}
\vskip -0.2in
\label{table:noise-exp-full}
\begin{center}
\begin{tiny}
\begin{sc}
\begin{tabular}{lrrrrrrrrr}
\toprule
 &\multicolumn{3}{c}{MNIST} &\multicolumn{3}{c}{Fashion-MNIST} &\multicolumn{3}{c}{Kuzushiji-MNIST} \\
 \cmidrule(lr){2-4}
 \cmidrule(lr){5-7}
 \cmidrule(lr){8-10}
\hspace{1em}  &$\lambda=0.1$ &$\lambda=0.2$ &$\lambda=0.5$ &$\lambda=0.1$ &$\lambda=0.2$ &$\lambda=0.5$ &$\lambda=0.1$ &$\lambda=0.2$ &$\lambda=0.5$ \\
\midrule
URE-GA &22.8$\pm$ 2.0 &21.1$\pm$ 4.4 &21.4$\pm$ 1.6 &20.2$\pm$ 6.7 &23.5$\pm$ 3.9 &22.6$\pm$ 3.1 &16.8$\pm$ 2.1 &16.4$\pm$ 2.8 &15.2$\pm$ 2.2\\
SCL &25.6$\pm$ 13.8 &23.9$\pm$ 10.3 &23.7$\pm$ 4.3 &23.9$\pm$ 7.8 &24.5$\pm$ 5.2 &26.0$\pm$ 3.2 &17.8$\pm$ 2.5 &17.8$\pm$ 3.2 &17.4$\pm$ 1.3\\
DM  &23.3$\pm$ 7.4 &22.4$\pm$ 8.7 &23.4$\pm$ 2.9 &24.1$\pm$ 7.1 &24.3$\pm$ 5.0 &25.6$\pm$ 3.9 &18.1$\pm$ 2.6 &17.6$\pm$ 2.4 &16.5$\pm$ 1.4\\
Fwd &91.1$\pm$ 0.7 &89.6$\pm$ 1.0 &82.5$\pm$ 3.6 &82.4$\pm$ 0.9 &81.4$\pm$ 0.9 &72.0$\pm$ 7.5 &\textbf{62.7$\pm$ 1.0} &60.9$\pm$ 0.9 &52.1$\pm$ 6.2\\
CPE-I  &75.7$\pm$ 2.0 &75.4$\pm$ 2.0 &73.8$\pm$ 2.2 &74.6$\pm$ 2.3 &73.9$\pm$ 2.2 &71.1$\pm$ 2.0 &47.0$\pm$ 1.4 &46.5$\pm$ 1.3 &43.4$\pm$ 1.1\\
CPE-F  &91.2$\pm$ 0.7 &90.2$\pm$ 1.0 &85.2$\pm$ 1.7 &82.2$\pm$ 1.2 &81.0$\pm$ 1.5 &75.4$\pm$ 3.3 &61.9$\pm$ 0.9 &\textbf{61.1$\pm$ 2.2} &\textbf{53.4$\pm$ 1.5}\\
CPE-T  &\textbf{91.3$\pm$ 0.7} &\textbf{90.5$\pm$ 0.8} &\textbf{85.7$\pm$ 1.6} &\textbf{82.6$\pm$ 1.3} &\textbf{81.6$\pm$ 1.3} &\textbf{78.0$\pm$ 1.6} &62.2$\pm$ 0.8 &\textbf{61.7$\pm$ 1.7} &\textbf{55.0$\pm$ 1.1}\\
\bottomrule
\end{tabular}
\end{sc}
\end{tiny}
\end{center}
\end{table*}

\newpage
\paragraph{Comparison of validation processes}
\begin{table}[t]
\caption{Comparison of \textbf{CPE-T}'s testing accuracies using different validation procedures.}
\label{table:comp-scel-ure-cpet}
\vskip -0.15in
\begin{center}
\begin{tiny}
\begin{sc}
\begin{tabular}{lrrrrrrrrr}
\toprule
&\multicolumn{3}{c}{MNIST} &\multicolumn{3}{c}{Fashion-MNIST} &\multicolumn{3}{c}{Kuzushiji-MNIST} \\
 \cmidrule(lr){2-4}
 \cmidrule(lr){5-7}
 \cmidrule(lr){8-10}
&Unif. &Weak &Strong &Unif. &Weak &Strong &Unif. &Weak &Strong\\
\midrule
\textbf\textit{linear} \\
\hspace{1em} URE   &90.3$\pm$ 0.6 &90.4$\pm$ 0.3 &91.8$\pm$ 0.5 &\textbf{82.1$\pm$ 0.3} &81.5$\pm$ 1.2 &82.6$\pm$ 1.3 &59.9$\pm$ 0.4 &60.0$\pm$ 0.9 &62.5$\pm$ 0.5\\
\hspace{1em} SCEL  &\textbf{90.5$\pm$ 0.2} &\textbf{90.6$\pm$ 0.1} &91.8$\pm$ 0.4 &82.0$\pm$ 0.3 &\textbf{82.1$\pm$ 0.5} &\textbf{83.2$\pm$ 1.2} &\textbf{60.3$\pm$ 0.5} &\textbf{60.6$\pm$ 0.5} &\textbf{63.0$\pm$ 0.3}\\
\textbf\textit{mlp} \\
\hspace{1em} URE   &92.7$\pm$ 0.5 &91.8$\pm$ 0.7 &90.4$\pm$ 6.5 &82.9$\pm$ 0.1 &83.0$\pm$ 0.3 &84.3$\pm$ 1.5 &\textbf{63.8$\pm$ 0.7} &63.8$\pm$ 1.9 &\textbf{74.5$\pm$ 2.7}\\
\hspace{1em} SCEL  &\textbf{92.8$\pm$ 0.6} &\textbf{92.1$\pm$ 0.2} &\textbf{95.2$\pm$ 0.5} &\textbf{83.0$\pm$ 0.1} &83.0$\pm$ 0.3 &\textbf{85.8$\pm$ 0.3} &63.6$\pm$ 0.4 &\textbf{64.6$\pm$ 0.4} &74.2$\pm$ 2.8\\
\bottomrule
\toprule
&$\lambda=0.1$ &$\lambda=0.2$ &$\lambda=0.5$ &$\lambda=0.1$ &$\lambda=0.2$ &$\lambda=0.5$ &$\lambda=0.1$ &$\lambda=0.2$ &$\lambda=0.5$ \\
\midrule
\textbf\textit{linear} \\
\hspace{1em} URE   &90.9$\pm$ 1.0 &90.2$\pm$ 0.8 &\textbf{86.1$\pm$ 1.3} &82.2$\pm$ 1.3 &81.2$\pm$ 1.4 &77.1$\pm$ 1.8 &\textbf{62.3$\pm$ 0.8} &60.6$\pm$ 0.9 &\textbf{55.3$\pm$ 2.3}\\
\hspace{1em} SCEL  &\textbf{91.3$\pm$ 0.7} &\textbf{90.5$\pm$ 0.8} &85.7$\pm$ 1.6 &\textbf{82.6$\pm$ 1.3} &\textbf{81.6$\pm$ 1.3} &\textbf{78.0$\pm$ 1.6} &62.2$\pm$ 0.8 &\textbf{61.7$\pm$ 1.7} &55.0$\pm$ 1.1\\
\textbf\textit{mlp} \\
\hspace{1em} URE   &83.7$\pm$ 9.7 &90.8$\pm$ 4.7 &82.9$\pm$ 9.4 &83.0$\pm$ 3.2 &74.8$\pm$ 10.1 &74.3$\pm$ 10.1 &68.5$\pm$ 11.4 &67.1$\pm$ 7.7 &57.2$\pm$ 16.3\\
\hspace{1em} SCEL  &\textbf{94.4$\pm$ 0.5} &\textbf{93.7$\pm$ 0.5} &\textbf{89.6$\pm$ 0.9} &\textbf{84.1$\pm$ 0.8} &\textbf{83.2$\pm$ 1.1} &\textbf{78.9$\pm$ 2.0} &\textbf{76.1$\pm$ 1.3} &\textbf{73.9$\pm$ 1.6} &\textbf{64.2$\pm$ 1.2}\\
\bottomrule
\end{tabular}
\end{sc}
\end{tiny}
\end{center}
\end{table}
\begin{table}[t]
\caption{Comparison of \textbf{Fwd}'s testing accuracies using different validation procedures.}
\label{table:comp-scel-ure-fwd}
\vskip -0.4in
\begin{center}
\begin{tiny}
\begin{sc}
\begin{tabular}{lrrrrrrrrr}
\toprule
&\multicolumn{3}{c}{MNIST} &\multicolumn{3}{c}{Fashion-MNIST} &\multicolumn{3}{c}{Kuzushiji-MNIST} \\
 \cmidrule(lr){2-4}
 \cmidrule(lr){5-7}
 \cmidrule(lr){8-10}
&Unif. &Weak &Strong &Unif. &Weak &Strong &Unif. &Weak &Strong\\
\midrule
\textbf\textit{linear} \\
\hspace{1em} URE   &90.5$\pm$ 0.2 &90.6$\pm$ 0.4 &91.6$\pm$ 0.7 &82.0$\pm$ 0.4 &81.6$\pm$ 1.2 &83.4$\pm$ 0.7 &59.9$\pm$ 0.9 &60.4$\pm$ 0.9 &62.6$\pm$ 0.7\\
\hspace{1em} SCEL  &90.5$\pm$ 0.2 &\textbf{90.7$\pm$ 0.2} &\textbf{91.9$\pm$ 0.4} &\textbf{82.2$\pm$ 0.3} &\textbf{82.6$\pm$ 0.3} &\textbf{83.8$\pm$ 0.2} &\textbf{60.4$\pm$ 0.6} &\textbf{61.2$\pm$ 0.3} &\textbf{63.2$\pm$ 0.2}\\
\textbf\textit{mlp} \\
\hspace{1em} URE   &94.4$\pm$ 0.2 &91.9$\pm$ 0.3 &95.3$\pm$ 0.4 &82.6$\pm$ 0.6 &83.0$\pm$ 1.0 &85.5$\pm$ 0.3 &73.5$\pm$ 1.6 &63.1$\pm$ 2.6 &74.1$\pm$ 4.8\\
\hspace{1em} SCEL  &94.4$\pm$ 0.2 &\textbf{92.0$\pm$ 0.2} &\textbf{95.5$\pm$ 0.2} &\textbf{83.0$\pm$ 0.1} &\textbf{83.3$\pm$ 0.2} &\textbf{86.1$\pm$ 0.5} &73.5$\pm$ 1.6 &\textbf{64.8$\pm$ 0.5} &\textbf{75.3$\pm$ 2.6}\\
\bottomrule
\toprule
&$\lambda=0.1$ &$\lambda=0.2$ &$\lambda=0.5$ &$\lambda=0.1$ &$\lambda=0.2$ &$\lambda=0.5$ &$\lambda=0.1$ &$\lambda=0.2$ &$\lambda=0.5$ \\
\midrule
\textbf\textit{linear} \\
\hspace{1em} URE   &91.1$\pm$ 0.7 &89.6$\pm$ 1.0 &82.5$\pm$ 3.6 &82.4$\pm$ 0.9 &81.4$\pm$ 0.9 &72.0$\pm$ 7.5 &\textbf{62.7$\pm$ 1.0} &60.9$\pm$ 0.9 &52.1$\pm$ 6.2\\
\hspace{1em} SCEL  &\textbf{91.4$\pm$ 0.5} &\textbf{90.5$\pm$ 0.5} &\textbf{83.9$\pm$ 2.6} &\textbf{83.2$\pm$ 0.3} &\textbf{82.4$\pm$ 0.4} &\textbf{76.3$\pm$ 2.8} &62.5$\pm$ 0.9 &\textbf{62.5$\pm$ 1.6} &\textbf{55.6$\pm$ 2.0}\\
\textbf\textit{mlp} \\
\hspace{1em} URE   &88.3$\pm$ 8.7 &83.9$\pm$ 10.7 &71.6$\pm$ 18.4 &84.8$\pm$ 0.6 &80.2$\pm$ 6.2 &62.9$\pm$ 20.1 &72.8$\pm$ 5.6 &67.6$\pm$ 7.5 &54.7$\pm$ 12.4\\
\hspace{1em} SCEL  &\textbf{94.4$\pm$ 0.3} &\textbf{93.5$\pm$ 0.3} &\textbf{84.5$\pm$ 4.1} &\textbf{85.0$\pm$ 0.3} &\textbf{84.0$\pm$ 0.5} &\textbf{76.5$\pm$ 2.5} &\textbf{76.4$\pm$ 1.1} &\textbf{73.8$\pm$ 1.2} &\textbf{59.9$\pm$ 3.4}\\
\bottomrule
\end{tabular}
\end{sc}
\end{tiny}
\end{center}
\end{table}
Table \ref{table:comp-scel-ure-cpet} and \ref{table:comp-scel-ure-fwd} provide comparison of validation process using URE and the proposed SCEL. In Table \ref{table:comp-scel-ure-cpet}, we observe that SCEL selects better parameters in most cases. We also observe that when the transition matrix is inaccurate, the parameters selected by SCEL tends to be more stable, especially when the base models are mlp. This demonstrates the superiority of SCEL despite not being an unbiased estimator of the classification accuracies. In Table \ref{table:comp-scel-ure-fwd}, we further apply SCEL to Fwd. Similarly, we observe that SCEL selects better parameters in most cases. This suggests that the proposed validation procedure can not only be applied to CPE but also earlier approaches. It enables a more robust approach to validate earlier methods.

\ %
\begin{figure}[h]
\begin{center}
\centerline{\includegraphics[scale=0.40]{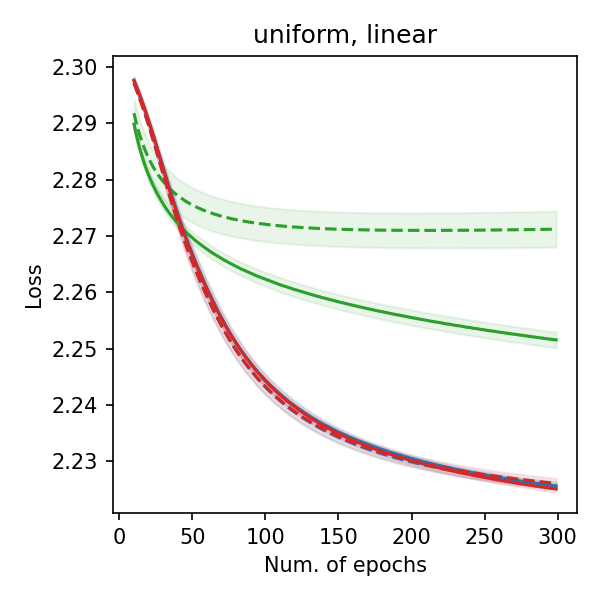}
\includegraphics[scale=0.40]{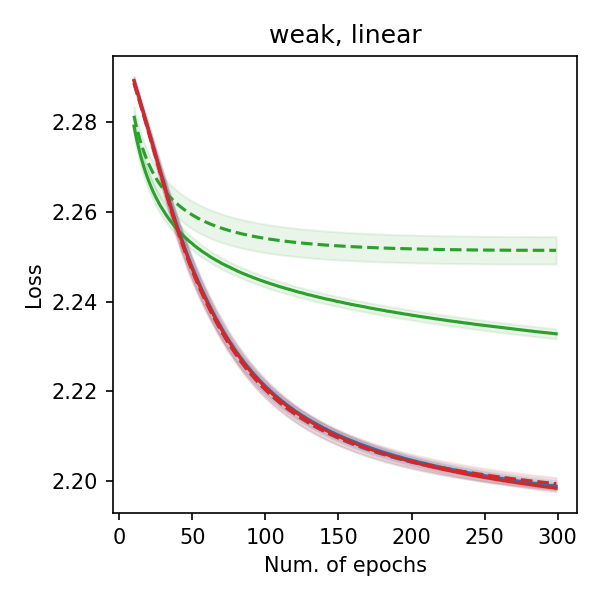}
\includegraphics[scale=0.40]{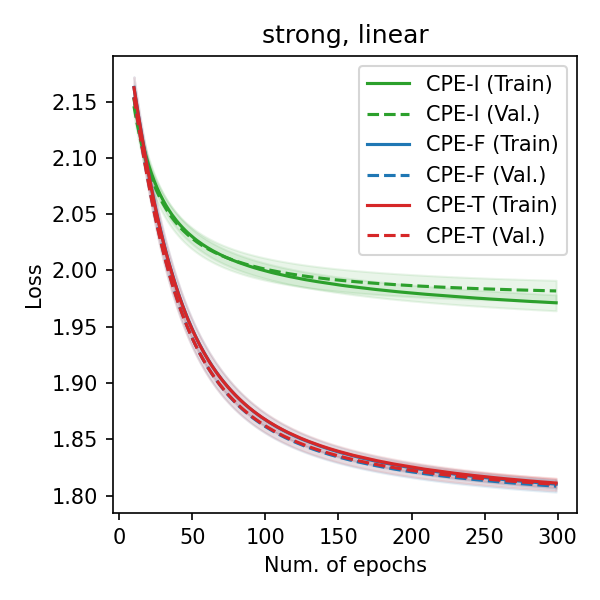}}
\centerline{\includegraphics[scale=0.40]{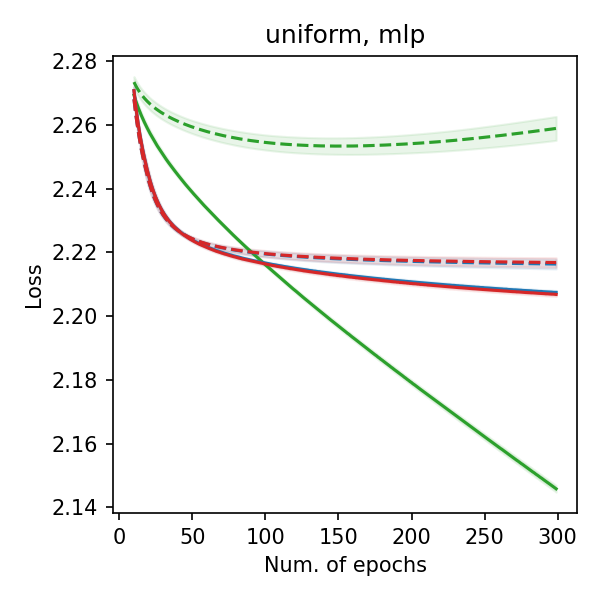}
\includegraphics[scale=0.40]{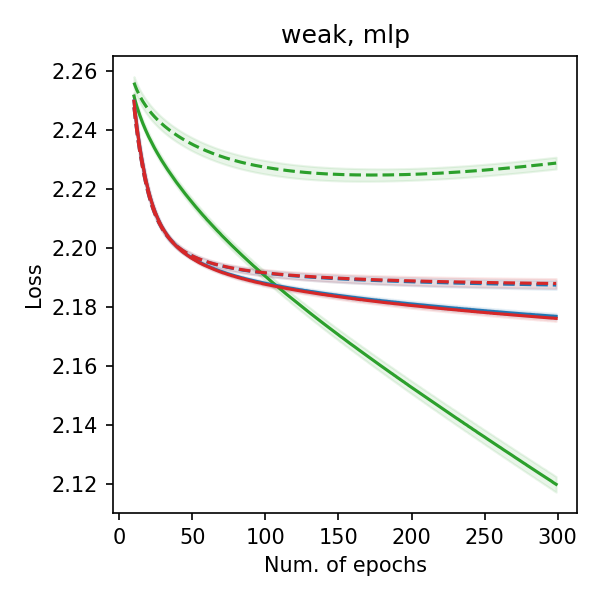}
\includegraphics[scale=0.40]{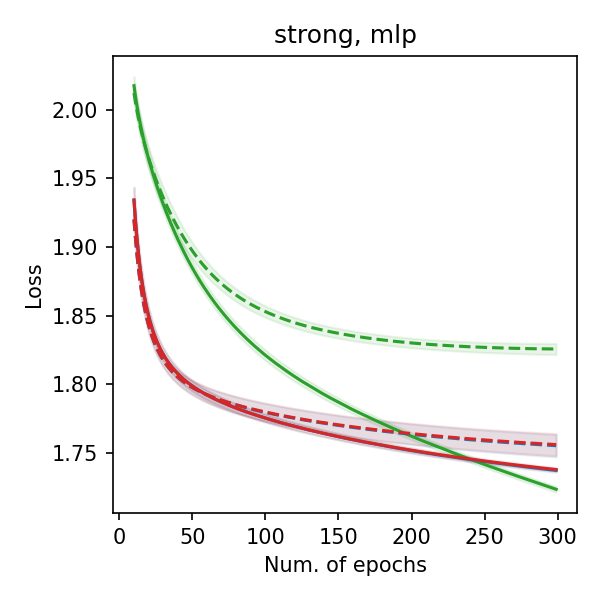}}
\end{center}
\vskip -0.25in
\caption{Comparison of the training and validation loss of CPE with different transition layers in MNIST under different transition matrices. CPE-F and CPE-T perform almost identically, so the red lines and blue lines overlap in the figures. The shaded area denotes the standard deviation of five random trials.}
\label{fig:train-val-loss}
\begin{center}
\centerline{\includegraphics[scale=0.40]{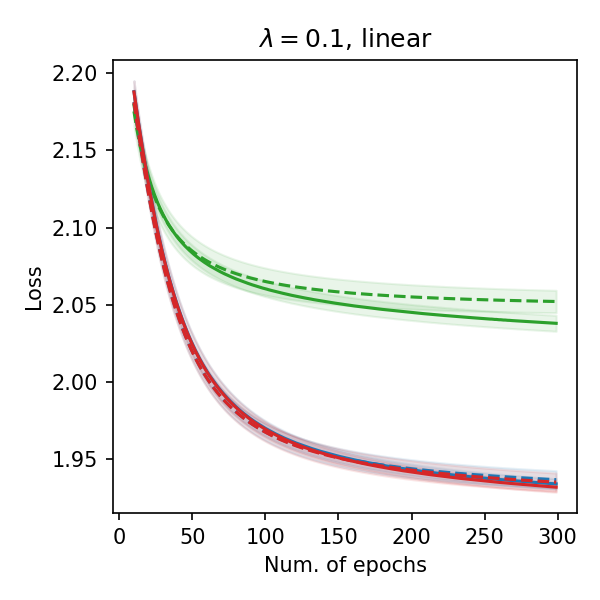}
\includegraphics[scale=0.40]{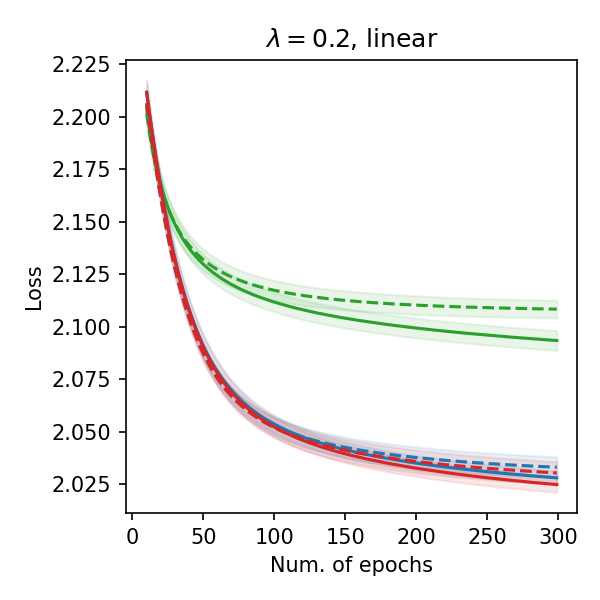}
\includegraphics[scale=0.40]{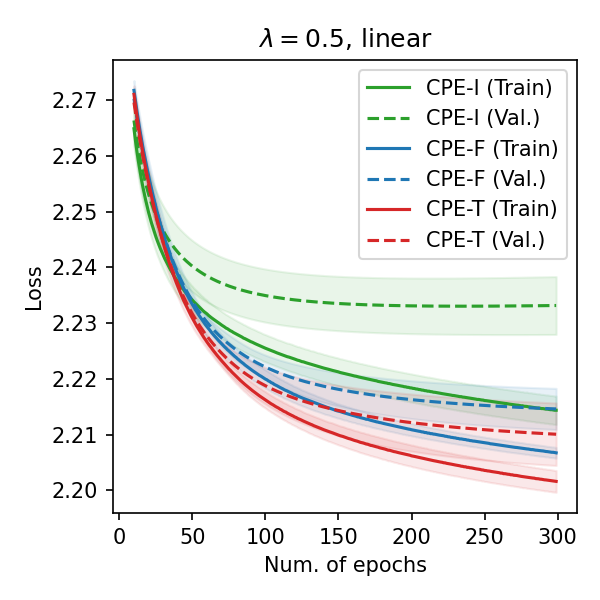}}
\centerline{\includegraphics[scale=0.40]{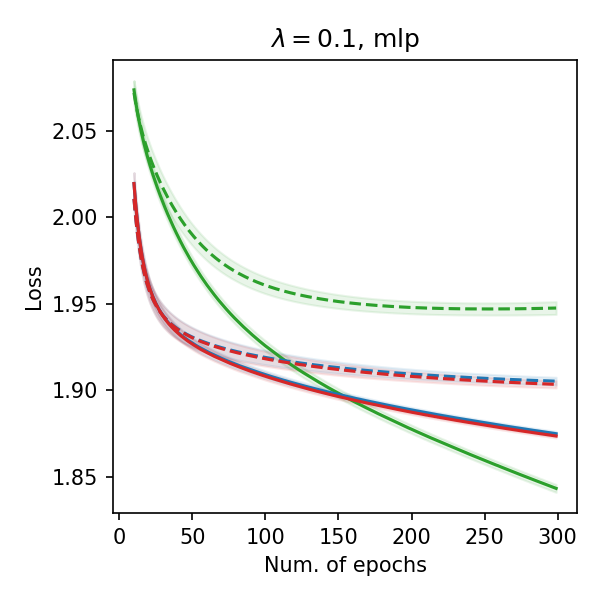}
\includegraphics[scale=0.40]{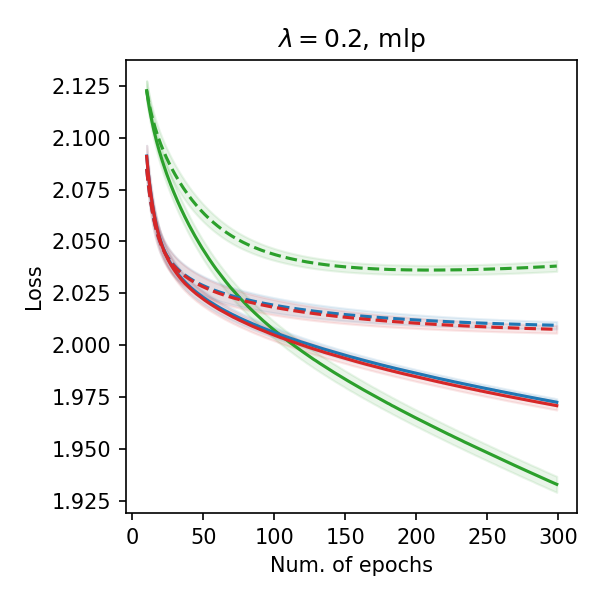}
\includegraphics[scale=0.40]{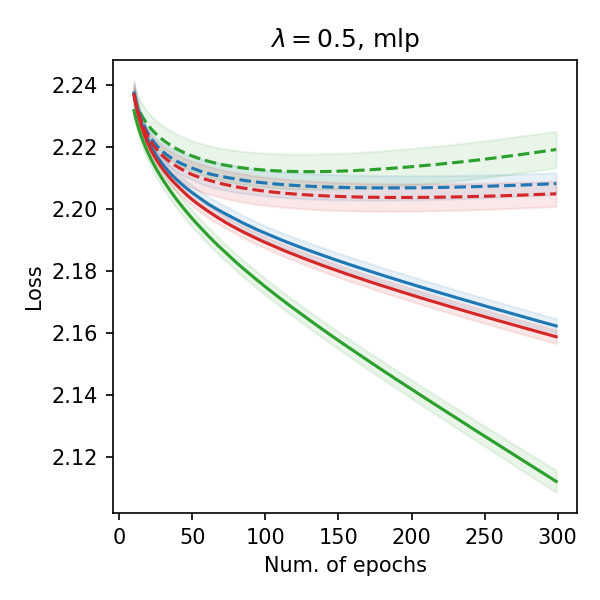}}
\end{center}
\vskip -0.25in
\caption{Comparison of the training and validation loss of CPE with different transition layers in MNIST under different noise level. CPE-F and CPE-T perform almost identically when $\lambda$ is small, so the red lines and blue lines overlap in those figures. The shaded area denotes the standard deviation of five random trials.}
\label{fig:train-val-loss-noise}
\end{figure}
\paragraph{Training and validation loss curves}
Figure \ref{fig:train-val-loss} and \ref{fig:train-val-loss-noise} demonstrate the loss curve of the proposed \textbf{CPE} framework.

\end{document}